\documentclass[runningheads]{llncs}

\usepackage{eccv}

\usepackage{eccvabbrv}

\usepackage{graphicx}
\usepackage{booktabs}

\usepackage{wrapfig}
\usepackage{multirow}
\usepackage{adjustbox}
\usepackage{colortbl}
\usepackage[table]{xcolor} 
\definecolor{1}{RGB}{255, 120, 120}   
\definecolor{2}{RGB}{255, 200, 120} 
\definecolor{3}{RGB}{255, 255, 120}  

\usepackage[accsupp]{axessibility}  
\usepackage{hyperref}

\usepackage{orcidlink}

\usepackage{paralist}
\renewcommand\paragraph[1]{\vspace{0.2cm}\noindent\textit{#1}}

\begin{document}

\title{GAINS: Gaussian-based Inverse Rendering from Sparse Multi-View Captures} 

\author{Patrick Noras\inst{1,2}\and
Jun Myeong Choi\inst{3}\and
Didier Stricker\inst{1,2}\and 
Pieter Peers\inst{4}\and
Roni Sengupta\inst{3}}

\authorrunning{P.~Noras et al.}

\institute{University of Kaiserslautern-Landau, Kaiserslautern, Germany \and
German Research Center for Artificial Intelligence, Kaiserslautern, Germany \and
University of North Carolina at Chapel Hill, NC, USA \and
College of William \& Mary, VA, USA}

{
\maketitle
\centering
\includegraphics[width=0.95\linewidth]{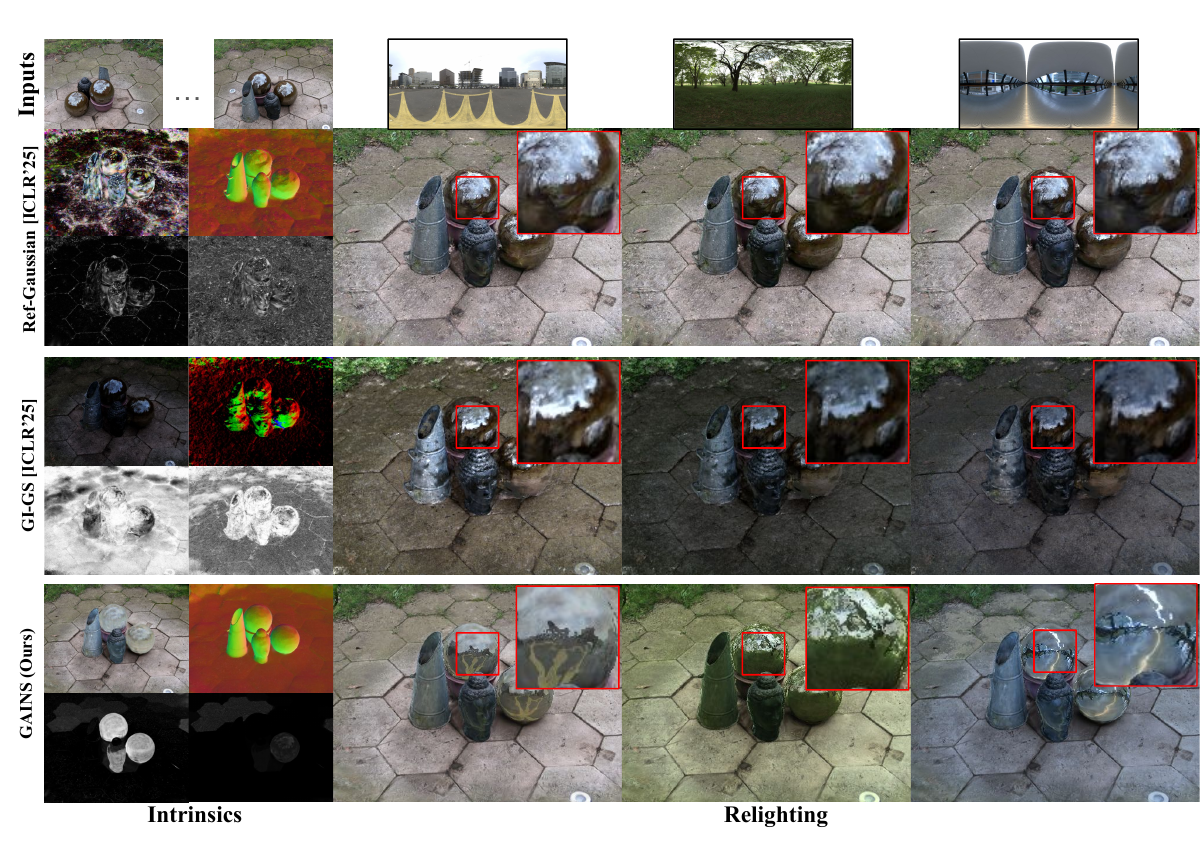}
\vspace{-1em}
\captionof{figure}{We introduce \textbf{GAINS}, GAussian-based INverse rendering from Sparse multi-view captures, which synergizes foundation models for monocular depth/normal, segmentation, intrinsic image decomposition (IID), and diffusion as priors, to better disambiguate reflectance from lighting, leading to better intrinsics, novel view synthesis and relighting compared to existing state-of-the-art approaches such as Ref-Gaussian~\cite{yao2025refGS} and GI-GS~\cite{chen2025gigs}. Prior methods often overfit diffuse (\eg, missing yellow reflection from the ground in the first relighting example for GI-GS) and/or specular reflections (\eg, for both Ref-Gaussian and GI-GS the reflected details remain unchanged under different lighting conditions). In contrast, \textbf{GAINS} improves estimation of material properties, leading to better relighting in novel views.}
\label{fig:teaser}
\vspace{-0.8em}
}

\begin{abstract}
%
Recent advances in Gaussian Splatting-based inverse rendering extend Gaussian primitives with shading parameters and physically grounded light transport, enabling high-quality material
recovery from dense multi-view captures. However, the accuracy of these methods degrades under sparse-view settings, where limited observations lead to severe ambiguity between geometry, reflectance, and lighting. We introduce GAINS (Gaussian-based Inverse rendering from Sparse multi-view captures), a two-stage inverse rendering framework that leverages foundation models as priors to stabilize geometry and material estimation. The core technical contribution of this paper is an inverse rendering framework that unifies foundation model priors with physically-based representations in an optimization scheme.
GAINS first refines geometry using monocular depth, normal, and diffusion priors, and then employs segmentation, intrinsic image decomposition (IID), and diffusion priors to regularize material recovery. Extensive experiments on synthetic and real-world datasets show that GAINS significantly improves material parameter accuracy, relighting quality, and novel-view synthesis compared to state-of-the-art Gaussian-based inverse rendering methods. While GAINS outperforms and remains competitive across a wide range of objects captured with 4 to 32 cameras, the improvement is particularly pronounced under sparse-view settings, where ambiguity is high and learning-based priors become especially beneficial. Project page: \href{https://patrickbail.github.io/gains/}{https://patrickbail.github.io/gains/}

\end{abstract}    
\section{Introduction}
\label{sec:intro}
Inverse rendering (IR) aims to recover the intrinsic 3D properties of a scene (\ie, geometry, material properties, and lighting) from multi-view images. IR serves as a key step toward physically grounded 3D scene understanding with many downstream applications such as novel-view synthesis, relighting, material and shape editing, etc. Over the past decades, inverse rendering has seen remarkable progress, evolving from early methods \cite{senguptaSfSNetLearningShape2018,Yu_2019_CVPR,sengupta2019neural} that use simple intrinsic representations such as surface normals, albedo, and spherical harmonics lighting, to modern approaches that employ rich and realistic intrinsic representations such as 3D Gaussian primitives, physically-based BRDFs, and realistic illumination \cite{liu2023nero,wang2024nep,hasselgren2022nvdiffrecmc,zhang2023neilf++,sun2023neuralpbir,gu2024IRGS,li2020inverse,lichy2021shape}. Simultaneously, the scope of inverse rendering has broadened from simple, controlled objects like faces to complex real-world scenes containing intricate geometry and diverse materials.

Despite these advances, scaling inverse rendering from simple objects to complex scenes with intricate details typically requires dense multi-view captures. Dense observations provide strong geometric and photometric constraints that help disambiguate material properties, lighting, and shape, all of which are entangled in the image formation process. However, when viewpoints are sparse, these constraints weaken, leading to overfitting and degraded performance. In such settings, existing methods often fail to recover accurate intrinsic properties, producing inconsistent material and lighting estimates. Moreover, traditional smoothness priors are inadequate for resolving the reflectance-lighting ambiguity, resulting in poor novel-view synthesis (NVS) and relighting.

To address these challenges, we propose GAINS (GAussian-based INverse rendering from Sparse multi-view captures), a novel inverse rendering framework designed to operate robustly under sparse-view settings. We focus on a Gaussian splatting-based framework since state-of-the-art approaches \cite{shi2025gir, yao2025refGS, chen2025gigs} found Gaussian splatting to be more effective for inverse rendering both in terms of quality and efficiency. GAINS follows the standard two-stage inverse rendering pipeline, where we first estimate geometry (Stage I), followed by an estimation of material properties and lighting (Stage II). The key technical contribution of this work lies in developing an inverse rendering framework that integrates physically-based modeling with multiple foundation models serving as priors within an optimization pipeline.

Geometry estimation forms the foundation of physically-based rendering, and thus, high-quality geometry is crucial for accurate material and lighting recovery. In Stage I, we leverage learning-based priors from \emph{monocular depth and normal prediction} networks as well as a \emph{latent diffusion model}, drawing inspiration from recent progress in sparse-view Gaussian-based reconstruction methods \cite{yang2024gaussianobject, xiong2023sparsegs}, as detailed in~\autoref{sec:geo_rec}. Each of these priors contributes towards improving geometric estimation accuracy, which further improves novel-view synthesis, BRDF estimation, and relighting.
In Stage II,  we introduce three complementary learning-based priors for improved material estimation:
(1) A \emph{segmentation guidance} that enforces multi-view consistency and reduces noise in material maps. However, segmentation guidance lacks intrinsic knowledge of surface reflectance, limiting its generalization to novel views and lighting conditions; (2) An \emph{Intrinsic Image Decomposition (IID) prior}, implemented using a state-of-the-art intrinsic image decomposition network, that provides a strong initialization for albedo, at the cost of poor cross-view consistency and weak generalization; and (3) a \emph{latent diffusion prior} that offers strong generalization capabilities. However, the latent diffusion prior struggles with material consistency and multi-view consistency.   Only by combining all three complementary priors (\autoref{sec:mat}), GAINS recovers robust material properties and lighting while achieving stable novel-view synthesis and relighting, even under sparse input conditions, as shown in ~\autoref{fig:teaser}.

We conduct extensive experiments on two synthetic benchmarks (Shiny Blen\-der \cite{verbin2022refnerf}, extended with ground-truth albedo and relighting, and Synthetic4Relight \cite{zhang2022invrender}, which additionally includes ground-truth roughness), as well as one real-world dataset~\cite{verbin2022refnerf}. GAINS consistently outperforms state-of-the-art Gaussian-based inverse rendering methods across geometry and material estimation tasks by a significant margin. For example, GAINS outperforms state-of-the-art Ref-Gaussian \cite{yao2025refGS} in relighting on Synthetic4Relight \cite{zhang2022invrender} by 28.16 vs. 24.29 in PSNR and on Shiny Blender\cite{verbin2022refnerf} by 22.29 vs. 17.26 in PSNR. While the improvements are especially pronounced under sparse-view conditions, our method also achieves competitive results with dense inputs. We further perform detailed ablation studies to analyze the contributions of each foundation model prior and demonstrate how their combination yields the most robust and physically consistent inverse rendering results.
\begin{figure*}[t]
  \centering
  \includegraphics[width=1\linewidth]{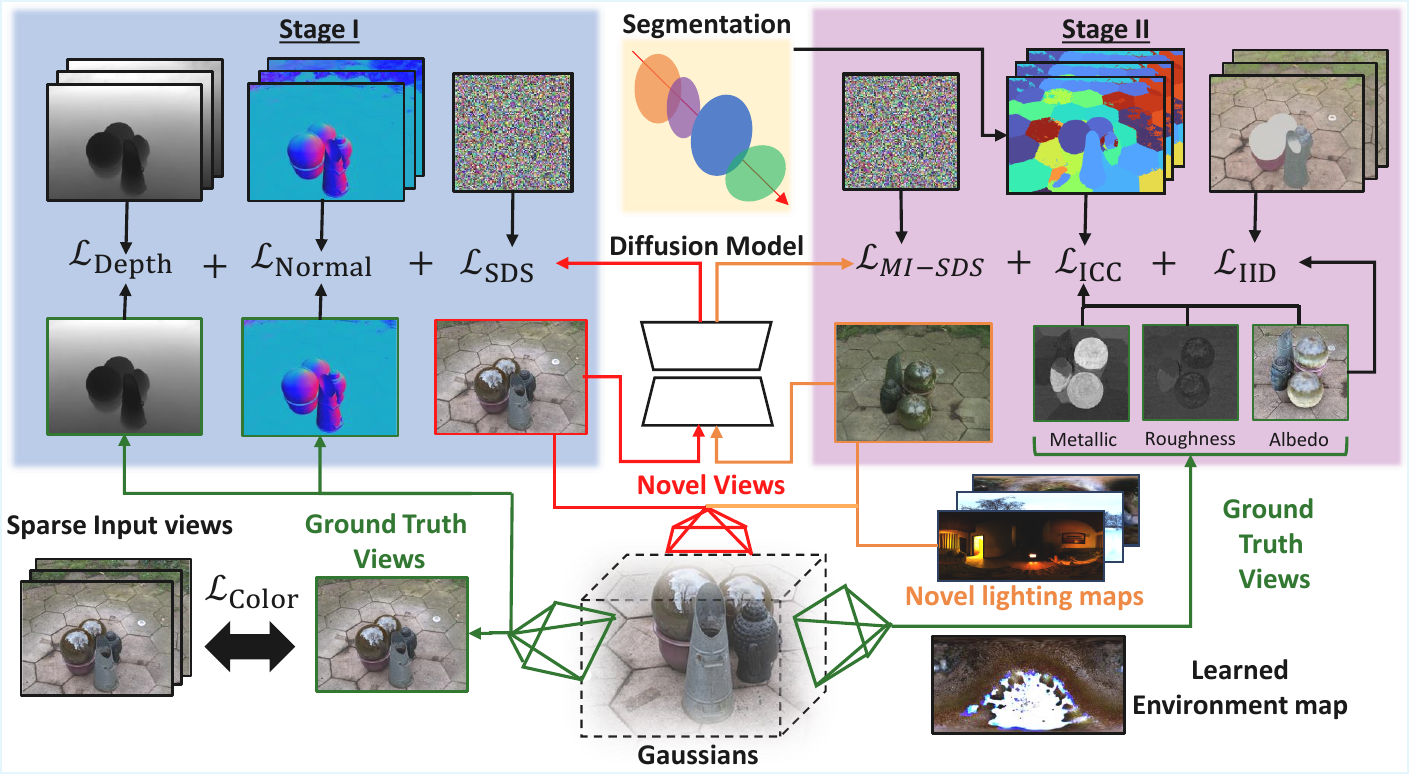}
  \vspace{-0.3cm}
  \caption{
  \textbf{GAINS} follows a two-stage inverse rendering pipeline. 
  In Stage I, we enhance geometry using learning-based priors from monocular depth, normal, and diffusion predictors. In Stage II, we introduce three complementary priors: segmentation, intrinsic image decomposition (IID), and diffusion, to improve material estimation, novel-view synthesis, and relighting. Each prior provides distinct benefits: segmentation boosts cross-view consistency of specular parameters but degrades albedo; IID improves albedo accuracy but remains view-inconsistent; diffusion enhances generalization to novel view and relighting but lacks material consistency. \textbf{GAINS} integrates these priors to achieve stable, high-quality material properties, leading to better relighting under novel views.
  }
  \vspace{-1.1em}
  \label{fig:pipeline}
\end{figure*}

\section{Related Work}
\label{sec:related_work}

\noindent \textbf{Inverse Rendering} (IR) decomposes images into geometry, material properties, and illumination. Classical methods estimate surface shape and spatially varying material properties from controlled captures using analytic BRDF models and global illumination optimization~\cite{sato1997object, debevec2004estimating, xia2016recovering}.
With the advent of learning-based scene representations such as NeRF~\cite{mildenhall2020nerf}, neural IR frameworks~\cite{zhang2021nerfactor, Jin2023TensoIR, srinivasan2021nerv, zeng2023relighting, wu2024gani} integrate simplified BRDF and lighting models to recover intrinsic scene parameters from multi-view inputs. Notably, NeRFactor~\cite{zhang2021nerfactor} estimates shape, materials, and lighting under unknown illumination using smoothness and BRDF priors, while GaNI~\cite{wu2024gani} and related work jointly optimize geometry and material parameters with near-field lighting and neural radiance caching.

\noindent \textbf{Inverse Rendering with Gaussian Splatting.} Recently, 3D Gaussian Splatting (3DGS)~\cite{kerbl3Dgaussians} emerged as an efficient explicit radiance representation, which subsequently has been extended to IR~\cite{jiang2023gaussianshader, yao2025refGS, chen2025gigs, shi2025gir, liang2024gs, R3DG2023}.
GaussianShader~\cite{jiang2023gaussianshader} augments Gaussian primitives with BRDF parameters, while Ref-Gaussian~\cite{yao2025refGS} and GI-GS~\cite{chen2025gigs} integrate physically-based rendering (PBR) with deferred shading and ray tracing. 
While Gaussian splatting based IR frameworks provide higher efficiency and fidelity than neural implicit representations, they assume dense multi-view captures and/or object-centric settings, making them less practical for real-world sparse-view scenarios.

\noindent \textbf{Inverse Rendering with Sparse Views.}
Earlier approaches for sparse-view or single-view IR rely on learned priors from synthetic data \cite{senguptaSfSNetLearningShape2018,sengupta2019neural,li2020inverse} without explicit geometric and light-transport modeling leading to poor generaliation to complex real world scenes. Recent works like RelitLRM~\cite{zhang2024relitlrm} leverage Large Reconstruction Models (LRMs) and generative priors for object relighting, but they can not estimate intrinsic components (\eg, material properties), nor do they generalize to complex real-world scenes.

Recent approaches for shape reconstruction from sparse views using NeRF- or 3DGS-based representations increasingly leverage learning-based priors, such as monocular depth, normal, or diffusion-based guidance~\cite{wang2023sparsenerf, xiong2023sparsegs, zhu2023FSGS, yang2024gaussianobject, huang2025fatesgs}, demonstrating that combining learned priors with explicit geometric modeling leads to significant performance gains. However, a sparse-input inverse rendering framework that jointly integrates such learning-based priors with explicit shape representations and physically based rendering remains largely unexplored.
\section{Overview}
Our goal is to reconstruct robust and accurate geometry and material properties from a sparse set of captured 
RGB images $\{I_i\}^N_{i=1}$ (with $N$ as low as 4-16 views) under known camera intrinsics and extrinsics $C_i = \{K_i, R_i, t_i\}$.
Our pipeline builds on 2D Gaussian Splatting (2DGS)~\cite{Huang2DGS2024}, and adapts it for robust Inverse Rendering (IR) from sparse input views.  We opt for 2DGS over 3D Gaussian Splatting~\cite{kerbl3Dgaussians} due to its more accurate geometry reconstruction which is essential for accurate material property estimation.  

\noindent \textbf{Scene representation.}
As in 2DGS, we parameterize the geometry as: $\mathcal{G} = \{\mu_j, S_j, R_j, \sigma_j\}^M_{j=1}$, where $\mu_j$ describes the 3D position of each primitive, along with opacity $\sigma_j$, scaling vector $S_j$ and rotation matrix $R_j$; the surface normal $n_j$ is also determined by $R_j$.

\noindent \textbf{Material and lighting representation.}
Following previous Gaussian splatting-based IR methods~\cite{yao2025refGS, shi2025gir, chen2025gigs}, we use a physically-based rendering (PBR) deferred shading approach. We employ a principled BRDF model~\cite{burley2012physically} and store for each Gaussian primitive the corresponding BRDF parameters: $\mathcal{M} = \{\boldsymbol{a_j}, r_j, m_j\}^M_{j=1}$, where $\boldsymbol{a_j} \in [0,1]^3$ is the albedo, $r_j \in [0,1]$ the roughness and $m_j \in [0,1]$ the metallicity.  For efficiency, we employ the split-sum approximation~\cite{karis2013real} to model the diffuse and specular surface reflectance:
\begin{eqnarray}
    L_{\text{diffuse}} &=& (1 - m) \cdot a \int_\Omega L(\omega_i)\frac{\omega_i \cdot n}{\pi}d\omega_i \\
    L_{\text{specular}}  &\approx& \underbrace{\int_\Omega \frac{DFG}{4(\omega_o \cdot n)}\omega_id\omega_i}_{\text{precomp. BRDF LUT}} \cdot \underbrace{\int_\Omega D \ L(\omega_i)d\omega_i}_{\text{prefiltered Env. Map}},
\end{eqnarray}
where $L(\omega_i)$ is the direct incident lighting stored as a $128 \times 128$ environment cubemap, $n$ the surface normal. $D$ the GGX microfacet distribution, $F$ Fresnel term, and $G$ geometric shadowing term. 

\noindent \textbf{Optimization.}
The goal of IR is to optimize shape $\mathcal{G}$, material reflectance $\mathcal{M}$, and lighting $L$ of the scene to minimize a re-rendering loss with respect to the input images.  Similarly to prior work, we solve the optimization in two stages: (1) optimize for shape $\mathcal{G}$, followed by (2) a joint estimation of material parameters $\mathcal{M}$ and lighting $L$. To improve robustness of the two-stage pipeline under sparse input views, we augment each stage with appropriate additional loss terms. We improve geometry estimation (\autoref{sec:geo_rec}) by leveraging additional learning-based priors from a latent diffusion model and a monocular depth and normal estimation model following established best practices from recent sparse-view Gaussian-based geometry reconstruction methods~\cite{yang2024gaussianobject, xiong2023sparsegs, huang2025fatesgs}.  We stabilize material and lighting estimation using three complementary learning-based priors: segmentation (\autoref{sec:segmentation}), IID (Intrinsic Image Decomposition;~\autoref{sec:iid}), and diffusion priors (\autoref{sec:sds}).  
\section{GAINS Stage I: Shape Estimation}
\label{sec:geo_rec}
We start with the standard 2DGS shape estimation loss, originally formulated for dense input views: $
%
    \mathcal{L} = \mathcal{L}_{\text{col}} + \lambda_{DD}\mathcal{L}_{DD} + \lambda_{NC}\mathcal{L}_{NC},
%
$ where, $\mathcal{L}_{\text{col}} = (1-\lambda)\mathcal{L}_1 + \lambda\mathcal{L}_{\text{D-SSIM}}$ is a re-rendering loss; $\mathcal{L}_{DD}$ is an optional depth distortion loss, reducing depth ambiguity by bringing intersected Gaussians closer along each ray, and $\mathcal{L}_{NC} = 1-{N_i}^T\tilde{N}_i$ is the normal consistency loss with $\tilde{N_i}$ being surface normals obtained from depth.

When applied to sparse input views, the above loss tends to overfit to the captured viewpoints, resulting in significant novel viewpoint artifacts. To better guide the IR optimization towards a robust and generalizable solution, we introduce additional learning based priors: (1) depth and normal guidance, and (2) diffusion guidance.

\noindent \textbf{Depth and Normal Guidance}
While learning-based depth priors have proven effective for geometry reconstruction~\cite{xiong2023sparsegs,huang2025fatesgs, zhu2023FSGS, yang2024gaussianobject}, their role in IR remains unclear, as hallucinations or systematic biases in monocular depth estimation may adversely affect material and lighting estimation. We include a depth loss by enforcing similarity to a monocularly estimated depth $\hat{D}_i$~\cite{ke2023repurposing}: $\mathcal{L}_{DC} = 1 - PCC(D_i, \hat{D_i})$, where $PCC(\cdot)$ is the Pearson Correlation Coefficient. Additionally, we also add a local depth ranking loss $\mathcal{L}_{DR}$~\cite{huang2025fatesgs} to prevent geometric collapse from hard constraints and alleviate long-range ambiguity. 
Moreover, given the importance of accurate surface normals in material estimation, we impose
normal smoothness via a total variation loss and an L1 loss on rendered normals $N_i$ and surface reconstructed normals $\hat{N}_i$. The reference normals $\tilde{N_i}$ are obtained from a monocular normal estimator~\cite{ke2023repurposing}. Together we obtain the normal guided loss $\mathcal{L}_N = \mathcal{L}_1(\tilde{N}_i, \hat{N}_i) + 0.1 \cdot\mathcal{L}_1(\tilde{N}_i, N_i) + 
TV(\tilde{N_i}, \hat{N}_i)$. As shown in the ablation study (\autoref{tab:ablation_stage1}), incorporating normal guidance improves high-frequency geometric fidelity and leads to more stable material estimation and rendering quality.

\noindent \textbf{Diffusion Guidance}
Inspired by recent successes in leveraging diffusion models for zero-shot 3D reconstruction~\cite{yi2023gaussiandreamer, liu2023zero1to3}, we guide the shape reconstruction through Score Distillation Sampling (SDS). Unlike depth and normal guidance, which provide pixel-aligned geometric constraints, the diffusion prior regularizes the reconstruction in unseen viewpoints $C_j$ by encouraging rendered images to lie on the manifold of the objects. Concretely, We sample $100$ novel viewpoints for which we render and perform a forward diffusion, yielding noisy latents: $Z_j = \alpha_t \mathcal{R}_{geo}(C_j, \mathcal{G}) + \sigma_t \epsilon$, where timestep $t \sim \mathcal{U}(0.02, 0.98)$ and noise $\epsilon \sim \mathcal{N}(\boldsymbol{0}, I_0)$. The SDS loss is then calculated as: $\mathcal{L}_{SDS} = \mathbb{E}_{t,\epsilon}\Big[w(t)||(\epsilon_{\phi}(Z_j;t) - \epsilon)||^2_2\Big]$. However, the SDS loss is not effective during early stages of the optimization when the shape reconstruction is far from the target shape. We therefore only include the SDS loss after iteration $10000$ (out of $16000$). With a guidance scale of $100$, the diffusion prior primarily suppresses high-frequency artifacts; because the geometry is mostly converged, hallucinations are limited.

\noindent \textbf{Outlier Removal}
If ground-truth alpha masks are provided, we additionally employ an Outlier Removal (OR) strategy. Here, a Binary-Cross Entropy (BCE) loss on reconstructed and ground truth alpha mask is employed, along with a KNN-based floater removal mechanism at every 1000 iterations. If no alpha masks are provided, we assume the scene to reconstruct the background as well.

\noindent\textbf{Summary}
The total loss function for shape estimation in Stage I is:
\vspace{-0.15cm}
    \begin{align}
        \mathcal{L}_{col} + \lambda_{DC} \mathcal{L}_{DC} + \lambda_{DR} \mathcal{L}_{DR} + \lambda_{NC} \mathcal{L}_{NC}  \nonumber
		+ \lambda_{N} \mathcal{L}_{N} + \lambda_{SDS} \mathcal{L}_{SDS} + \lambda_{BCE} \mathcal{L}_{BCE},
        \vspace{-0.5cm}
	\end{align}
with $\lambda_{DC} = 0.005$, $\lambda_{DR} = 10$, $\lambda_{NC} = 1$, $\lambda_{N} = 0.25$, $\lambda_{SDS} = 0.0001$, $\lambda_{BCE} = 0.75$. The complete Stage I pipeline is summarized in~\autoref{fig:pipeline} (left). All hyperparameters are fixed without any scene-specific tuning or prior-fine-tuning.
\section{GAINS Stage II: Material Properties}
\label{sec:mat}
In the second stage, we jointly optimize material properties $\mathcal{M} = \{(a_j, r_j, m_j)\}$ and environment lighting $L$ using the re-rendering loss $\mathcal{L}_{\text{color}}$ over $7000$ additional iterations. However, multiple combinations of different $\mathcal{M}$ and $L$ produce similar appearances, leading to an ill-posed and ambiguous material estimation.
This problem is further exacerbated when only a sparse set of input views is available. Although rendering under the recovered lighting and material properties appears visually plausible, the material properties are overfitted, noisy, inconsistent, and not suitable for relighting (\autoref{fig:teaser}).
To alleviate the impact of overfitting under sparse inputs, we complement the loss with additional learning based priors to better guide the optimization towards more plausible material properties, and thus resulting in more robust relighting. We synergize three complementary priors: segmentation (\autoref{sec:segmentation}), Intrinsic Image Decomposition (IID) (\autoref{sec:iid}), and diffusion guidance (\autoref{sec:sds}).

\vspace{-0.5cm}
\subsection{Segmentation Guidance}
\label{sec:segmentation}

While diffuse albedo varies significantly due to texture variations, we observe that specular material properties are typically consistent within semantically similar regions of an object or scene.  This suggests that segmentation cues can help guide the estimation of specular material properties. 
As a single semantic object can contain multiple subparts with distinct materials, we employ subpart segmentation to define a consistency loss. Instead of constraining the material parameters to be identical for each subpart, we minimize their variance to support mixed materials and to account for imperfect segmentation boundaries.

To include segmentation guidance, we extend the Gaussian primitives with a one-hot vector $E_j = \{0,1\}^K$ that encodes each Gaussian's class membership. We employ training-free segmentation mask lifting~\cite{chacko2025lifting} by rendering SH colored images from $V=100$ sampled novel view points orbiting the scenery. For each image $\{\hat{I_v}\}^V_{v=1}$ we generate (potentially view-inconsistent) segmentation masks with SAM~\cite{ravi2024sam2} along with mask-related features $\{f_{v,s} = [g_{v,s}, h_{v,s}]\}^V_{v=1}$ where $f_{v,s}$ is the feature vector for a mask region $s$ of view $v$ and $g_{v,s}, h_{v,s}$ are  CLIP~\cite{radford2021learning} and DINOv2~\cite{oquab2023dinov2} features respectively. For each view, we lift the associated mask and features to the Gaussian that contributes most ($\alpha$-blending wise) and merge already collected Gaussian objects from previous lifting operations with geometric and feature similarity scores.

Next, we leverage the possibly imprecise segmentation masks to define an intra-class consistency (ICC) on the roughness and metallicity parameters:
\vspace{-0.25cm}
\begin{align}
    \mathcal{L}_{ICC} &= \frac{1}{|S_i|} \sum_{s_i \in S_i} \gamma(|s_i|) \cdot \Big( \mathcal{L}_{r,m} + \mathcal{L}_a +\mathcal{L}_e \Big) \\
    \mathcal{L}_{r,m} &= Var(R_{s_i}) + Var(M_{s_i}) \\
    \mathcal{L}_a &= \Big((1-R_{s_i}) + M_{s_i}\Big) \cdot \frac{1}{3} \sum^3_{c=1} Var(A_{s_i}^{(c)}) \\
    \mathcal{L}_{e} &= \Big( (1-M_{s_i}) + R_{s_i} \Big) \cdot E_{s_i}
\end{align}
where $S_i$ is the set of all masks rendered from a captured viewpoint $i$, $A_{s_i}$, $R_{s_i}$, and $M_{s_i}$ are the rendered masked albedo, roughness, and metallicity, along with $E_{s_i}$ representing the masked re-render loss. $\gamma(|s_i|) = \exp(25 \cdot\frac{|I_i||s_i|}{|S_i|^2})$ is a scaling function depending on the size of the mask $s_i$. The first variance reduction term $\mathcal{L}_{r,m}$ reduces irregularities for specular roughness and metallicity across all masks.  While in general we attribute texture variations to the diffuse texture, an ambiguity exists in the case of mirror-like materials that reflects texture from the environment lighting. To address this ambiguity, we bias the optimization to attribute texture details to specular reflections in the case of mirror-like materials by adding a loss term $\mathcal{L}_{a}$ that encourages a variance reduction in diffuse albedo in regions where we observe low roughness and high metallicity. Lastly, regions with high specularity typically suffer from larger geometry errors, and thus higher material estimation errors. We therefore add a loss $\mathcal{L}_e$ to encourage more specular in such areas, albeit with a low weight to avoid attributing all errors to specular.

\vspace{-0.5cm}
\subsection{Intrinsic Image Decomposition Prior}
\label{sec:iid}
While segmentation improves consistency in roughness and metallicity over multiple viewpoints, it is not suited for diffuse albedo which often contains high-frequency texture variations that are difficult to estimate from sparse viewpoints. In the absence of appropriate regularization, the model will overfit diffuse albedo and fail to disentangle lighting effects from intrinsic albedo, which ultimately leads to a subpar relighting performance.

Recent advances in monocular Intrinsic Image Decomposition (IID)~\cite{ke2023repurposing, zeng2024rgb, Sartor:2025:TCD} enable high quality diffuse albedo decompositions, albeit potentially view-inconsistent. We therefore propose to leverage the IID to regularize the diffuse albedo estimation while accounting for potential view-inconsistencies: $
%
    \mathcal{L}_{IID} = \beta(\tau) \cdot \mathcal{L}_2(A_i, \hat{A_i}),
%
$ where $\hat{A_i}$ is the IID diffuse albedo for the captured image $I_i$ obtained using Teamwork~\cite{Sartor:2025:TCD}. For outdoor data, we opt for RGB2X~\cite{zeng2024rgb}, as we found it to produce more reliable decompositions on outdoor scenes used in our experiments. To mitigate the influence of view-inconsistency in the IID, we weight the loss by $\beta(\tau)$ (a linear decrease) which depends on the current iteration step $\tau$, thereby relaxing the influence of the IID diffuse albedo as the model converges.  We opt against using additional IID supervision for specular roughness and metallicity as we found these parameters to be less robust and exhibit more inconsistencies across views.

\vspace{-0.5cm}
\subsection{Multi-illuminated Score Distillation Sampling}
\label{sec:sds}
While both segmentation guidance and IID guidance improve the accuracy of the specular and diffuse material properties respectively, they do not necessarily provide good generalization to novel viewpoints or lighting.
To address these issues, we take inspiration from our usage of SDS in the first stage to aid in reducing unrealistic artifacts on novel viewpoints, and introduce a modified SDS loss to improve view and lighting-consistency for material properties.
To avoid overfitting to the learned environment map $L$, we render relit images $\mathcal{R}_{mat}(C_j, \mathcal{G}, \mathcal{M}, \mathcal{E}_l)$ from a novel view $C_j$ under randomly selected lighting from a predetermined set $\{\mathcal{E}_l\}^{\mathcal{|E}|}_{l=1}$, and define the SDS loss as: $
%
	\mathcal{L}_{MI-SDS} = \mathbb{E}_{t,\epsilon}\Big[w(t)||(\epsilon_{\phi}(Z_j^l;t) - \epsilon)||^2_2\Big],
%
$ where $Z_j^l = \alpha_t \mathcal{R}_{mat}(C_j, \mathcal{G}, \mathcal{M}, \mathcal{E}_l) + \sigma_t \epsilon$ is the noisy diffusion latent of the rendered scene under novel view and lighting.
Similarly as in the geometry reconstruction stage, we start the diffusion guidance once the optimization solution has stabilized (\ie, after $3000$ steps). As in Stage I, with a guidance scale of $100$, the diffusion prior mainly suppresses high-frequency artifacts 
, introducing only limited hallucinated details.

\vspace{-0.5cm}
\subsection{Final Loss}

Each loss contributes to a specific enhancement of the material estimation process: parameter accuracy, multi-view consistency, or novel views and lighting generalization. We combine all losses as:
\vspace{-0.15cm} 
\begin{align*}
    \mathcal{L}_{mat} &= \mathcal{L}_{color} + \lambda_{ICC} \mathcal{L}_{ICC} + \lambda_{IID} \mathcal{L}_{IID} 
    + \lambda_{MI-SDS} \mathcal{L}_{MI-SDS} + \lambda_{TV}\mathcal{L}_{TV},
   \vspace{-1.0cm} 
\end{align*}
where $\lambda_{ICC} = 0.1$, $\lambda_{IID} = 2$, $\lambda_{MI-SDS} = 0.0001$ and $\lambda_{TV} = 0.1$. $\mathcal{L}_{TV}$ represents a total variation loss on our reconstructed lighting, acting as a smoothing term. \autoref{fig:pipeline} (right) summarizes the full Stage II pipeline. Reported hyperparameters are again fixed without any scene-specific tuning or prior-fine-tuning.
\begin{figure*}[!t]
  \centering
   \includegraphics[width=1\linewidth]{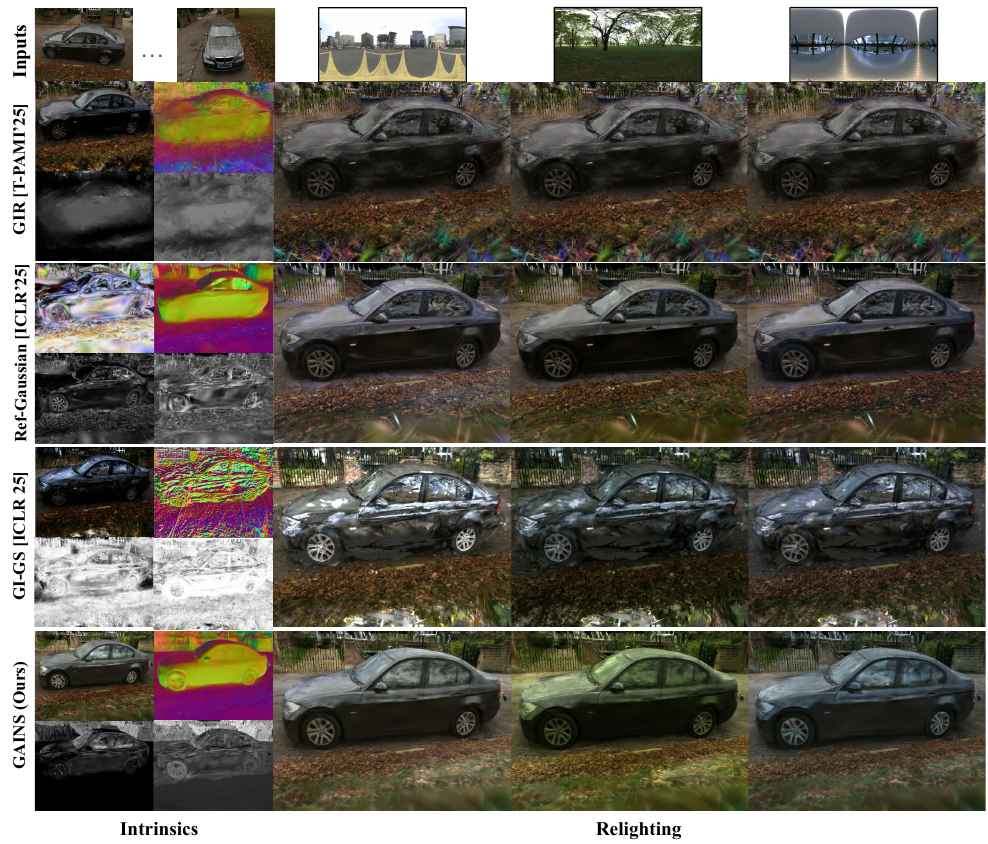}
   \vspace{-1em}
   \caption{\textbf{Qualitative comparison of intrinsic estimation and relighting on the \textit{sedan} scene from the Ref-Real dataset \cite{verbin2022refnerf} reconstructed from 8 views.} Column 1 shows novel-view intrinsic renderings in a 2×2 layout: (top) rendered albedo and surface normals, (bottom) specular roughness and metallicity. Columns 2–4 show relighting results under three different environment maps from novel viewpoints. GAINS recovers significantly more accurate intrinsics, enabling more realistic relighting.}
   \label{fig:main_result}
\end{figure*}

\section{Evaluation}
\noindent \textbf{Evaluation Framework.} We compare GAINS against state-of-the-art IR frameworks that use Gaussian Splatting for both objects and scenes: Ref-Gaussian~\cite{yao2025refGS}, GI-GS~\cite{chen2025gigs}, GShader~\cite{jiang2023gaussianshader}, R3DG~\cite{R3DG2023}, MaterialRefGS \cite{zhang2025materialrefgs} and SVG-IR \cite{SVGIR2025}. We evaluate on two synthetic datasets (\ie, the Shiny Blender dataset~\cite{verbin2022refnerf} modified to include albedo and relighting evaluation and without `ball' object; and Synthetic4Relight~\cite{zhang2022invrender} for NVS, albedo, relighting and roughness), and a real-world dataset (Ref-Real~\cite{verbin2022refnerf}), on which we quantitatively evaluate NVS performance only. We also evaluate GIR~\cite{shi2025gir} on real scenes from Ref-Real, but we found that it fails to reconstruct meaningful geometry and reflectance (\autoref{fig:main_result}) while requiring long per-scene training time (over 7~hours). Therefore, we opted not to include GIR in the evaluation over the synthetic datasets.
We evaluate NVS, albedo, and relighting using three metrics: PSNR, SSIM, and LPIPS, while for roughness we employ MSE. To account for the albedo-lighting ambiguity, we use scale invariant losses to evaluate the albedo for each approach by uniformly scaling each albedo map to minimize MSE w.r.t.~ground-truth before applying an error metric. All evaluations are performed on 8 sparse views uniformly sampled from the dense views.

\begin{table*}[t!]
\caption{\textbf{Quantitative evaluation on the synthetic Synthetic4Relight \cite{zhang2022invrender} dataset.} GAINS estimates better albedo and roughness then baselines leading to better relighting performance. Additionally, GAINS outperforms existing methods in novel-view synthesis.
(\colorbox{1}{\raisebox{0pt}[1ex][0ex]{Red}} = best, 
    \colorbox{2}{\raisebox{0pt}[1ex][0ex]{Orange}} = 2nd best, 
    and \colorbox{3}{\raisebox{0pt}[1ex][0ex]{Yellow}} = 3rd best)
  }  
\label{tab:errorcombined}
\vspace{-0.5em}
\centering
\begin{adjustbox}{max width=\textwidth,center}
\begin{tabular}{l | c c c | c c c | c c c | c | c}
 & \multicolumn{10}{c}{\textbf{Synthetic4Relight \cite{zhang2022invrender} dataset (8 views)}} \\
\hline
\multirow{2}{*}[-0.2em]{\textbf{Methods}} &
\multicolumn{3}{c|}{\textbf{NVS}} &
\multicolumn{3}{c|}{\textbf{Albedo}} &
\multicolumn{3}{c|}{\textbf{Relight}} &
\textbf{Roughness} &
\multirow{2}{*}[-0.2em]{\textbf{Runtime} $\downarrow$} \\
\cline{2-11}
 & \textbf{PSNR} $\uparrow$ & \textbf{SSIM} $\uparrow$ & \textbf{LPIPS} $\downarrow$ &
\textbf{PSNR} $\uparrow$ & \textbf{SSIM} $\uparrow$ & \textbf{LPIPS} $\downarrow$ &
 \textbf{PSNR} $\uparrow$ & \textbf{SSIM} $\uparrow$ & \textbf{LPIPS} $\downarrow$ &
 \textbf{MSE} $\downarrow$ \\
\hline
 GShader \cite{jiang2023gaussianshader} & 15.11 & 0.71 & 0.24 & - & - & - & 12.58 & 0.11 & 0.39 & 0.09 & $\approx$5h \\
 MaterialRefGS \cite{zhang2025materialrefgs} & 19.10 & 0.86 & 0.16 & 14.56 & 0.75 & 0.24 & 17.95 & 0.84 & 0.18 & 0.76 & $\approx$2h30m \\
 SVG-IR \cite{SVGIR2025} & 24.21 & 0.87 & \cellcolor{2}0.11 & 15.89 & \cellcolor{3}0.87 & \cellcolor{2}0.12 & \cellcolor{3}25.22 & \cellcolor{2}0.90 & \cellcolor{2}0.10 & \cellcolor{1}0.03 & $\approx$2h \\
 R3DG \cite{R3DG2023} & \cellcolor{3}26.92 & \cellcolor{3}0.90 & \cellcolor{2}0.11 & \cellcolor{2}22.15 & \cellcolor{2}0.88 & \cellcolor{2}0.12 & \cellcolor{2}27.42 & \cellcolor{2}0.90 & \cellcolor{2}0.10 & \cellcolor{2}0.04 & \cellcolor{3}$\approx$1h30m \\
 Ref-Gaussian \cite{yao2025refGS} & 24.05 & 0.83 & \cellcolor{2}0.11 & 17.37 & 0.83 & \cellcolor{3}0.17 & 24.29 & \cellcolor{3}0.85 & 0.13 & \cellcolor{2}0.04 & \cellcolor{1}$\approx$45m \\
 GI-GS \cite{chen2025gigs} & \cellcolor{2}27.04 & \cellcolor{2}0.91 & \cellcolor{3}0.12 & \cellcolor{3}20.47 & \cellcolor{2}0.88 & \cellcolor{2}0.12 & 23.92 & 0.82 & \cellcolor{3}0.11 & \cellcolor{3}0.05 & \cellcolor{3}$\approx$1h30m\\
 Ours & \cellcolor{1}29.32 & \cellcolor{1}0.94 & \cellcolor{1}0.08 & \cellcolor{1}22.34 & \cellcolor{1}0.91 & \cellcolor{1}0.10 & \cellcolor{1}28.16 & \cellcolor{1}0.93 & \cellcolor{1}0.09 & \cellcolor{2}0.04 & \cellcolor{2}$\approx$1h20m \\
 \hline
\end{tabular}
\end{adjustbox}
\vspace{1em}

\caption{\textbf{Quantitative evaluation on the Shiny Blender~\cite{verbin2022refnerf} dataset and real Ref-Real \cite{verbin2022refnerf} dataset.} On Shiny Blender, GAINS estimates better albedo and better relighting results compared to existing methods. On the Ref-real and Shiny Blender dataset, GAINS also performs better novel-view synthesis than existing approaches.}
\label{tab:tensorIR}
\vspace{-0.5em}
\centering
\small
\begin{adjustbox}{max width=\textwidth,center}
\begin{tabular}{l | c c c | c c c | c c c | c || c c c }
& \multicolumn{10}{c||}{\textbf{Shiny Blender \cite{verbin2022refnerf} dataset (8 views)}}
& \multicolumn{3}{c}{\textbf{Ref-Real \cite{verbin2022refnerf} dataset (8 views)}} \\
\hline
\multirow{2}{*}[-0.2em]{\textbf{Methods}} &
\multicolumn{3}{c|}{\textbf{NVS}} &
\multicolumn{3}{c|}{\textbf{Albedo}} &
\multicolumn{3}{c|}{\textbf{Relight}} &
\textbf{Normal} &
\multicolumn{3}{c}{\textbf{NVS}} \\
\cline{2-14}
 & \textbf{PSNR} $\uparrow$ & \textbf{SSIM} $\uparrow$ & \textbf{LPIPS} $\downarrow$ &
\textbf{PSNR} $\uparrow$ & \textbf{SSIM} $\uparrow$ & \textbf{LPIPS} $\downarrow$ &
\textbf{PSNR} $\uparrow$ & \textbf{SSIM} $\uparrow$ & \textbf{LPIPS} $\downarrow$ &
 \textbf{MAE} $\downarrow$ &
 \textbf{PSNR} $\uparrow$ & \textbf{SSIM} $\uparrow$ & \textbf{LPIPS} $\downarrow$ \\
\hline
 GShader \cite{jiang2023gaussianshader} & 10.73 & 0.52 & 0.35 & - & - & - & 6.22 & 0.13 & 0.53 & 55.14 & 15.26 & 0.38 & 0.46 \\
 R3DG\cite{R3DG2023} & \cellcolor{2}20.97 & \cellcolor{2}0.83 & \cellcolor{2}0.17 & \cellcolor{3}16.62 & \cellcolor{3}0.79 & \cellcolor{3}0.19 & 12.43 & \cellcolor{2}0.69 & \cellcolor{3}0.22 & \cellcolor{2}24.43 & 18.67 & 0.42 & 0.46 \\
 Ref-Gaussian \cite{yao2025refGS} & 18.60 & \cellcolor{3}0.80 & \cellcolor{3}0.18 & 15.82 & 0.77 & 0.22 & \cellcolor{2}17.26 & \cellcolor{3}0.66 & \cellcolor{2}0.20 & \cellcolor{3}25.54 &\cellcolor{3}19.39 & \cellcolor{2}0.45 & \cellcolor{2}0.36 \\
 GI-GS \cite{chen2025gigs} & \cellcolor{3}20.15 & 0.79 & 0.21 & \cellcolor{2}21.37 & \cellcolor{2}0.87 & \cellcolor{2}0.14 & \cellcolor{3}15.44 & 0.57 & 0.27 & 63.55 & \cellcolor{2}19.64 &\cellcolor{3}0.44 & \cellcolor{3}0.37 \\
 Ours & \cellcolor{1}23.90 & \cellcolor{1}0.89 & \cellcolor{1}0.12 & \cellcolor{1}21.73 & \cellcolor{1}0.89 & \cellcolor{1}0.12 & \cellcolor{1}22.29 & \cellcolor{1}0.89 & \cellcolor{1}0.12 & \cellcolor{1}11.64 & \cellcolor{1}20.28 & \cellcolor{1}0.56 & \cellcolor{1}0.34 \\
 \hline
\end{tabular}
\end{adjustbox}
\vspace{-1em}
\end{table*}

\begin{figure*}
  \centering
  \includegraphics[width=1\linewidth]{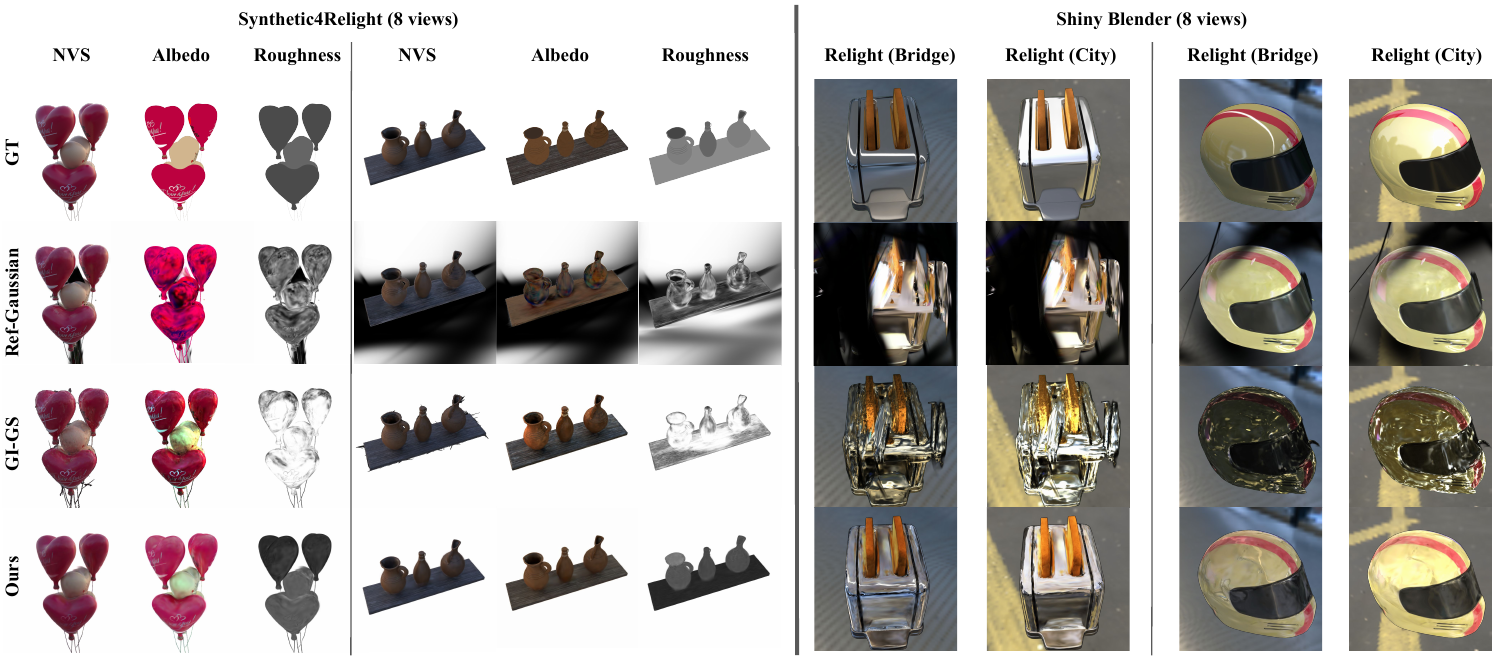}
  \vspace{-1em}
  \caption{\textbf{Qualitative comparison on Synthetic4Relight~\cite{zhang2022invrender} and Shiny Blender~\cite{verbin2022refnerf} trained with 8 views}. While all methods produce reasonable NVS, our method's estimates significantly better albedo, roughness and relighting than GI-GS and Ref-Gaussian that overfit to limited training views and fail to disentangle reflectance from lighting.}
  \label{fig:syn4s_exp}
\end{figure*}

\begin{figure*}[t!]
  \centering
  \includegraphics[width=1\linewidth]{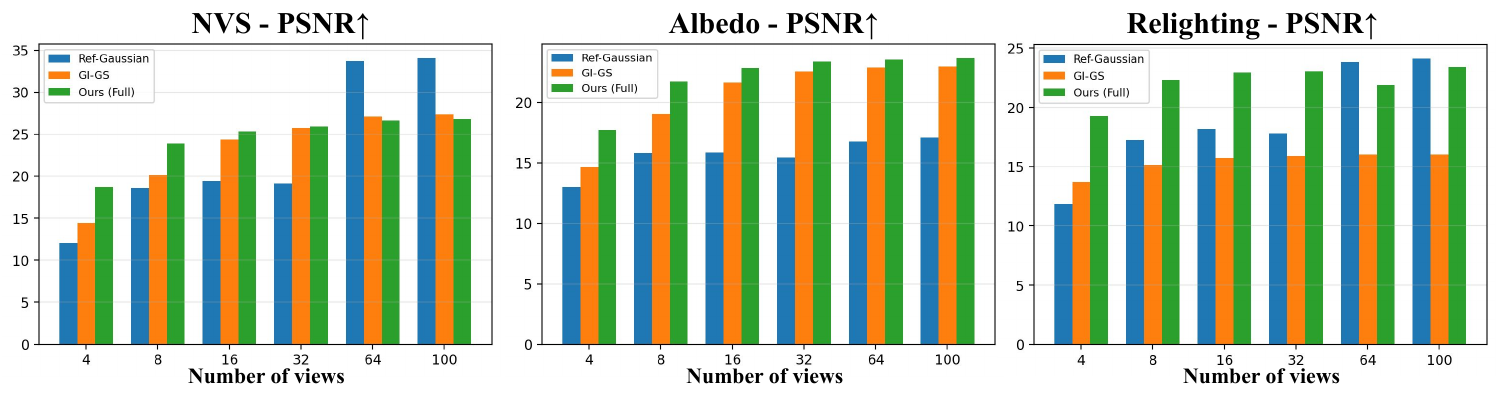}

  \vspace{-0.75em}
  
  \includegraphics[width=1\linewidth]{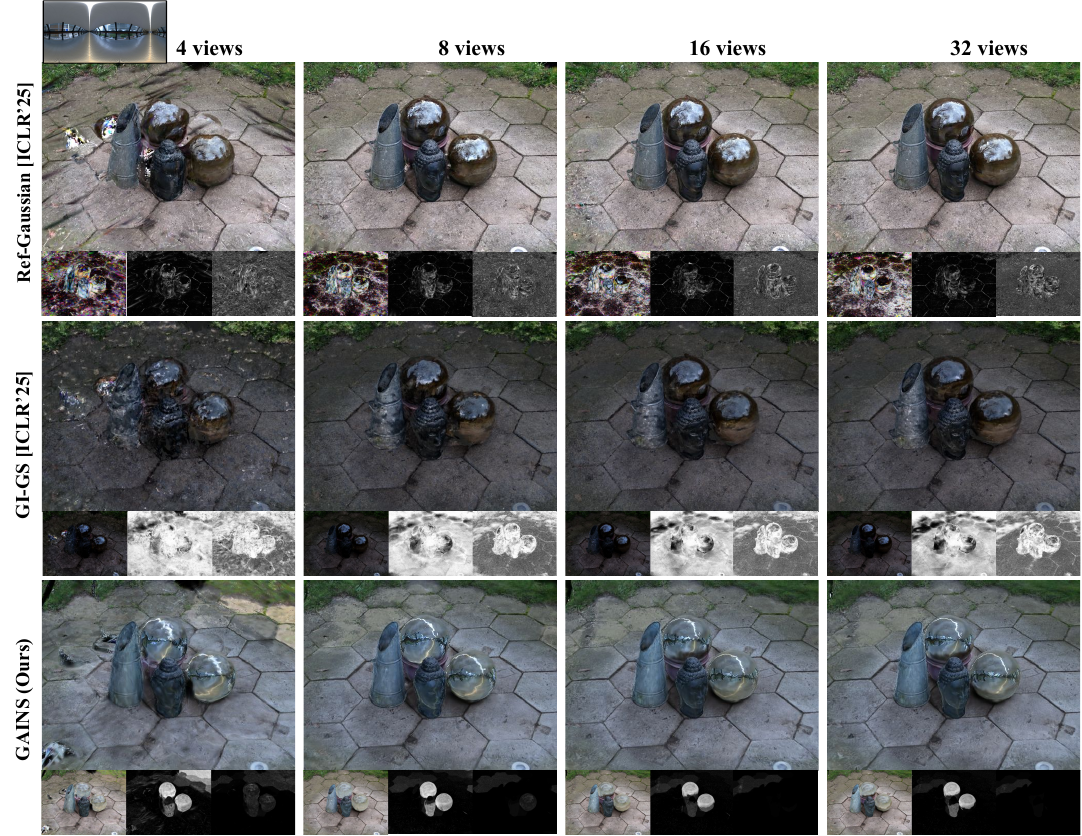}
  \vspace{-0.75em}
  \caption{\textbf{Comparison of our method with Ref-Gaussian~\cite{yao2025refGS} and GI-GS~\cite{chen2025gigs} for varying number of input views.} Top: Quantitative Evaluation on Shiny Blender~\cite{verbin2022refnerf}. Bottom: Qualitative comparison on the \textit{gardenspheres} scene from Ref-Real \cite{verbin2022refnerf} -- relit images on top and material maps at bottom. GAINS consistently produces robust relighting and material estimation from 4-32 views and remains competitive thereafter.}
  \label{fig:spheres_view_inc}
  \vspace{-1em}
\end{figure*}

\noindent \textbf{Performance Evaluation.} \autoref{tab:errorcombined}, and~\ref{tab:tensorIR} quantitatively summarize evaluation on the three datasets. In general, GAINS yields more accurate albedo maps and relighting across both Shiny Blender and Synthetic4Relight compared to all prior methods. Furthermore, GAINS achieves overall better results for NVS than all competing methods. However, error metrics do not always capture important visual differences. Therefore, we also provide qualitative comparisons in~\autoref{fig:teaser}, \ref{fig:main_result}, and~\ref{fig:syn4s_exp}. From the qualitative comparison, especially on the real-world Ref-Real dataset (\autoref{fig:teaser} and \ref{fig:main_result}), we note less baking of specular reflections, and generally more accurate relighting results, \eg, the reflections of the ground (missing for Ref-Gaussian) and the sky (baking of captured reflection for both Ref-Gaussian and GI-GS) on the ball in~\autoref{fig:teaser}. While GI-GS is the 2nd best method in NVS, we observe that it produces less accurate shape and normals (\eg, the noisy normals on the car in the sedan scene in~\autoref{fig:main_result}), indicating that the GI-GS NVS performance is mainly due to overfitting. As shown in~\autoref{fig:syn4s_exp}, our method demonstrates a clear advantage in material estimation under sparse-view settings. While Ref-Gaussian suffers from floaters and GI-GS typically recovers reasonable geometry but fails to produce accurate or stable albedo and roughness maps, our approach, supported by learning-based priors, delivers consistently superior NVS quality and more reliable albedo, roughness, and relighting estimations across all scenes.

\begin{wrapfigure}[14]{r}{0.55\textwidth}
    \centering
    \includegraphics[width=1\linewidth]{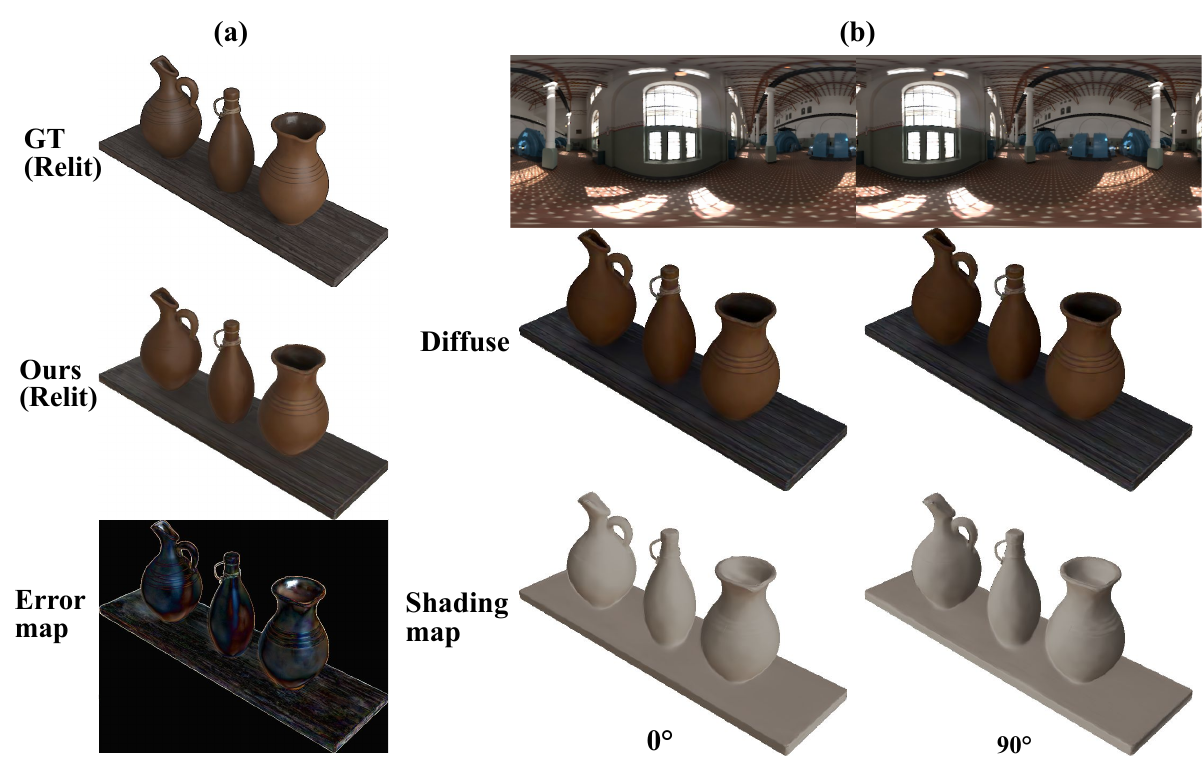}
   \caption{(a) Relighting comparison against GT with error map ($\times 5$). (b) Diffuse directional shading evidence.}
   \label{fig:dir_shading}
\end{wrapfigure}

\noindent \textbf{Evidence of Directional Relighting} In \autoref{fig:dir_shading}, we provide qualitative results under a novel environment map. Specifically, (a) compares our relit rendering against the ground truth along with the corresponding error map, while (b) visualizes diffuse directional shading evidence by rotating the environment map by $90^\circ$. These results highlight our method's ability to accurately reproduce both specular highlights and diffuse directional shading under novel illumination.

\noindent \textbf{Ablation: Number of input views.}
In addition to our comprehensive quantitative and qualitative evaluation on both synthetic and real datasets, we further analyze the behavior of our method as the number of available input views increases. \autoref{fig:spheres_view_inc} summarizes the quantitative comparison for PSNR on NVS, albedo and relighting on Shiny Blender~\cite{verbin2022refnerf}. We observe that in sparse-view scenarios (4-8 views), our method consistently outperforms all baselines across all evaluation scenarios by a significant margin. As the number of views increases (16-32), GI-GS competes in NVS; however, our method continues to produce competitive results within a small margin and continues to surpass in relighting. On denser view counts (64-100), Ref-Gaussian surpasses GAINS in NVS and begins to match our relighting performance. This improvement is largely due to its relaxed optimization objective: instead of explicitly optimizing albedo, it relies on the Stage~I SH representation and focuses on appearance fitting. Additionally, it continues scene densification during Stage~II, which further improves view synthesis quality as more observations become available. While all methods benefit from additional views, GAINS starts from a more stable and reliable baseline in the extremely sparse-view regime, demonstrating stronger robustness under challenging conditions.
Complementing our quantitative experiments, we also provide qualitative relighting and material estimation results (albedo, metallicity, and roughness) in~\autoref{fig:spheres_view_inc} on the \textit{gardenspheres} scene from the Ref-Real~\cite{verbin2022refnerf} dataset. Across all view counts, GAINS demonstrates strong and stable material reconstruction, producing accurate specular reflections and consequently superior relighting performance. Moreover, our roughness and metallic estimates remain stable and robust. This further demonstrates GAINS' advantage in low-view settings, where competing methods often suffer from noisy, irregular, or unstable material reconstructions. Improvements observed under dense inputs mainly stem from enhanced geometry estimation and improved lighting recovery. In contrast competing methods exhibit pronounced shape degradation in low-view settings (4-8 views).

\begin{table*}[t!]
\caption{\textbf{Stage I Ablation} with and without depth, normal, diffusion (SDS), and outlier removal (OR) on the Shiny Blender dataset~\cite{verbin2022refnerf} without Stage II components.}
\label{tab:ablation_stage1}
\vspace{-0.5em}
\centering
\begin{adjustbox}{max width=\textwidth,center}
\begin{tabular}{l | c c c | c c c | c c c | c}
\multicolumn{11}{c}{\textbf{Shiny Blender \cite{verbin2022refnerf} Stage I Ablation (8 views)}} \\
\hline
\multirow{2}{*}{\textbf{Method Variant}} &
\multicolumn{3}{c|}{\textbf{NVS}} &
\multicolumn{3}{c|}{\textbf{Albedo}} &
\multicolumn{3}{c|}{\textbf{Relight}} &
\textbf{Normal}\\
\cline{2-11}
 &
\textbf{PSNR} $\uparrow$ & \textbf{SSIM} $\uparrow$ & \textbf{LPIPS} $\downarrow$ &
\textbf{PSNR} $\uparrow$ & \textbf{SSIM} $\uparrow$ & \textbf{LPIPS} $\downarrow$ &
\textbf{PSNR} $\uparrow$ & \textbf{SSIM} $\uparrow$ & \textbf{LPIPS} $\downarrow$ &
\textbf{MAE} $\downarrow$ \\
\hline
Ours full & \cellcolor{1}23.67 & \cellcolor{1}0.88 & \cellcolor{1}0.12 & \cellcolor{1}21.18 & \cellcolor{1}0.88 & \cellcolor{1}0.14 & \cellcolor{1}21.96 & \cellcolor{1}0.86 & \cellcolor{1}0.15 & \cellcolor{1}11.64 \\
\hline
Ours w/o SDS & \cellcolor{2}23.57 & \cellcolor{1}0.88 & \cellcolor{2}0.13 & \cellcolor{2}21.09 & \cellcolor{2}0.87 & \cellcolor{1}0.14 & \cellcolor{2}21.91 & \cellcolor{1}0.86 & \cellcolor{1}0.15 & \cellcolor{2}11.70 \\
Ours w/o SDS, OR & \cellcolor{3}21.75 & \cellcolor{3}0.80 & \cellcolor{3}0.15 & \cellcolor{3}19.98 & \cellcolor{3}0.86 & \cellcolor{1}0.14 & \cellcolor{3}21.26 & \cellcolor{3}0.77 & \cellcolor{3}0.16 &\cellcolor{3}13.79 \\
Ours w/o Normal, SDS, OR & 19.42 & 0.79 & 0.16 & 18.79 & 0.82 & \cellcolor{1}0.14 & 19.41 & 0.74 & 0.17 & 19.38 \\
 \hline
Ours w/o Depth, Normal, SDS, OR & 16.73 & 0.71 & 0.21 & 16.41 & 0.81 & \cellcolor{2}0.17 & 16.71 & 0.67 & 0.21 & 32.23 \\
\hline
\hline
\end{tabular}
\end{adjustbox}
\vspace{-1em}
\end{table*}

\begin{figure*}[!t]

\begin{tabular}{cc}
\begin{adjustbox}{max width=0.5\textwidth}
\begin{tabular}{l | c | c | c }
\multicolumn{4}{c}{\textbf{Shiny Blender \cite{verbin2022refnerf} Stage II Ablation (8 views)}} \\
\hline
\multirow{2}{*}{\textbf{Method Variant}} &
\multicolumn{1}{c|}{\textbf{NVS}} &
\multicolumn{1}{c|}{\textbf{Albedo}} &
\multicolumn{1}{c}{\textbf{Relight}} \\
\cline{2-4}
 &
\textbf{PSNR} $\uparrow$ &
\textbf{PSNR} $\uparrow$ &
\textbf{PSNR} $\uparrow$ \\
\hline
Ours full & \cellcolor{1}23.90 & \cellcolor{2}21.73 & \cellcolor{1}22.29 \\
\hline
Ours w/o MI-SDS & \cellcolor{2}23.88 & \cellcolor{3}21.66 & \cellcolor{2}22.26 \\
Ours w/o Seg & 23.75 & \cellcolor{1}21.92 & \cellcolor{3}22.23 \\
Ours w/o IID & \cellcolor{3}23.84 & 21.38 & 22.07 \\
 \hline
Ours w/o Seg, IID, MI-SDS & 23.67 & 21.18 & 21.96\\
\hline
\hline
\end{tabular}
\end{adjustbox}

\begin{minipage}[t]{0.5\textwidth}
\vspace{-0.8cm}
\centering
\includegraphics[width=\textwidth]{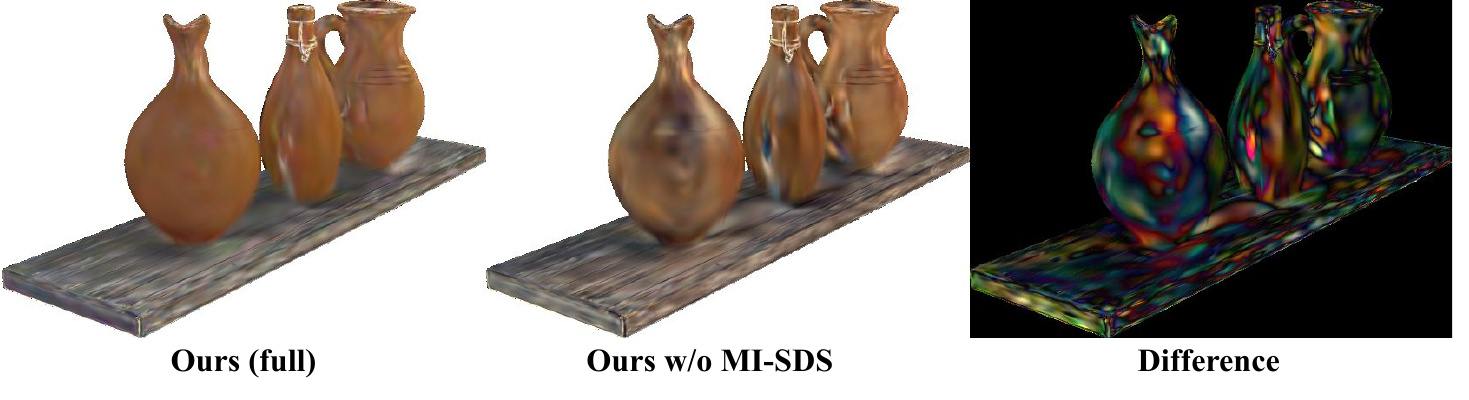}
\end{minipage}
\end{tabular}

\label{tab:ablation_stage2}

  \centering
  \includegraphics[width=1\linewidth]{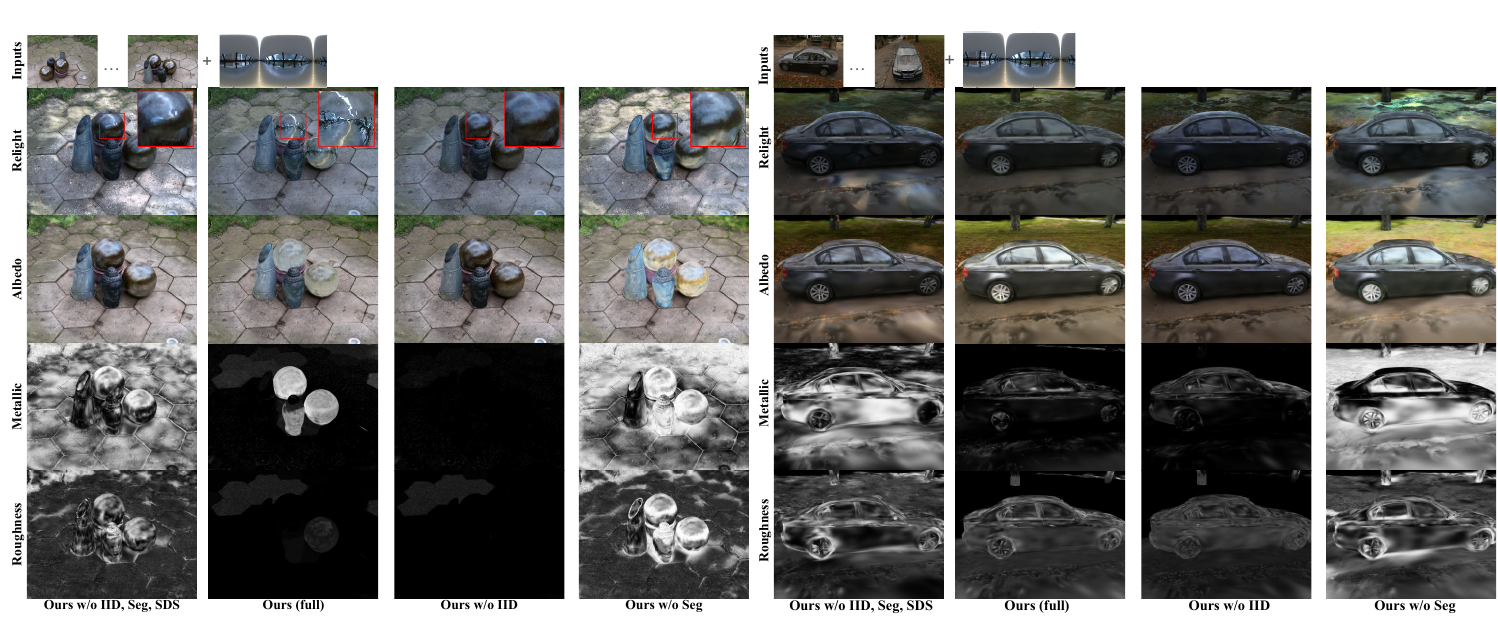}
  \caption{\textbf{Stage II Ablation.} Top left: Quantitative Evaluation on Shiny Blender~\cite{verbin2022refnerf}. Top right: Qualitative comparison of resulting albedo w/ and w/o MI-SDS with error map ($\times$5). Bottom: Qualitative comparison on the \textit{gardenspheres} and \textit{sedan} scenes from Ref-Real \cite{verbin2022refnerf}. We analyze three priors: intrinsic components (IID), segmentation (Seg), and diffusion (MI-SDS). Visual results on real data reveal their necessity - without IID material maps degrade and without segmentation multi-view consistency breaks. }
  \label{fig:prior_sing_exp}
  \vspace{-1em}
\end{figure*}

\noindent \textbf{Ablation: Learning-based priors in Stage I and Stage II.}
We conduct an ablation study on the Shiny Blender~\cite{verbin2022refnerf} dataset using 8 input views for both Stage I (\autoref{tab:ablation_stage1}) and Stage II (\autoref{fig:prior_sing_exp}). 
For Stage I, by distilling depth and normal priors into our shape learning stage, we are able to achieve robust normal and geometry reconstruction. Additionally, applying our outlier removal strategy for synthetic data, we effectively remove floaters outside the input view-space.  All components enable consistent improvement to achieve the best quality.
For Stage II (\autoref{fig:prior_sing_exp}), we observe that removing the IID prior reduces albedo reconstruction fidelity and leads to inconsistent material separation, resulting in degraded relighting performance. 
Excluding the segmentation prior produces incorrect object and material boundaries, yielding noisy or unstable specular components and harming novel-view synthesis. 
Finally, removing the diffusion prior leads to an overall drop in performance across all tasks, demonstrating its critical role in stabilizing optimization and improving generalization under sparse-view settings. More specifically, we notice a drop in reconstruction quality from unobserved views. Overall, the full model consistently achieves the strongest results across NVS, albedo, and relighting. %
%
%

\noindent\textbf{Limitations.}
First, GAINS ignores global illumination, which is more challenging for complete scenes than for isolated objects.  Second, while we enforce smoothness, the geometric normals can exhibit a slight waviness, which is especially noticeable on very specular objects.  As a consequence of wavy normals, some specularity can get baked into the diffuse albedo for very specular materials. Separating geometric and shading normals could potentially alleviate this.

\section{Conclusion}
\label{sec:conclusion}
We introduced GAINS, a Gaussian-based inverse rendering framework tailored for sparse-view settings. By combining segmentation, IID, and diffusion priors within a two-stage optimization pipeline, GAINS effectively stabilizes geometry and material estimation. Extensive experiments on synthetic and real-world datasets show that GAINS achieves improved relighting accuracy and novel-view synthesis compared to prior Gaussian-based IR methods. Qualitative results further demonstrate improved material–lighting separation and reduced reflection baking. GAINS highlights the power of synergizing complementary learning priors for physically consistent inverse rendering under sparse observations.
\section*{Acknowledgements}
Patrick Noras is supported by the DAAD-PROMOS scholarship during their research stay as a visiting scholar at the University of North Carolina at Chapel Hill. This work is partially supported by a National Institute of Health (NIH) NIBIB project \#R21EB035832 and \#R21EB037440 and National Science Foundation (NSF) CAREER Award \#2543161.

%
%
\bibliographystyle{splncs04}
\bibliography{main}

\pagebreak
\begin{center}
\textbf{\large GAINS: Gaussian-based Inverse Rendering from Sparse Multi-View Captures} \\
\vspace{0.5em}Supplemental Material
\end{center}
\appendix

\section{Overview of Appendices}
We categorize our appendices in the following way:
\begin{itemize}
    \item \autoref{sec:add_exp} provides additional details of our experimental setup, including the computational resources used and factors influencing the performance of our method.
    \item \autoref{sec:add_res} presents additional results on the TensorIR~\cite{Jin2023TensoIR} dataset, along with intermediate segmentation maps produced during training.
\end{itemize}

\section{Experiment, and Performance Details}
\label{sec:add_exp}
Unless stated otherwise, all experiments were conducted on a single NVIDIA RTX~A4500 GPU with 20\,GB of VRAM. Our method operates in linear color space; datasets provided in sRGB are internally linearized, and final predictions are converted back to sRGB before computing the corresponding loss terms. This ensures that all color-dependent computations remain physically consistent while metrics remain comparable to prior work.\\
Depth/normal and albedo models are used only as offline preprocessing ($\approx$5s and $\approx$30s for 8 views), which is negligible compared to the overall optimization time ($\approx$1h20m). Semantic processing with SAMv2, CLIP, and DINOv2 is a one-time intermediate step ($\approx$3-5min), independent of the number of training views. During optimization, only a single latent diffusion model is active for a part of the optimization ($\frac{1}{3}$ of Stage~I and $\frac{1}{2}$ of Stage~II). The computational overhead introduced by priors is negligible in sparse settings and scales only with the number of input images, while optimization runtime remains unchanged. Therefore, training efficiency depends only on two factors: the timing of the diffusion-prior activation and the number of views used during the intermediate segmentation-lifting stage. Notably, segmentation quality has an impact on the performance of the Gaussian representation. While using more views for the iterative segmentation merging improves consistency and reduces object-level ambiguities, it also increases computational overhead: $\approx$1min for 25 views and $\approx$5min for 100 views. This trade-off becomes particularly important in novel-view synthesis (NVS) and relighting tasks, where inaccurate or overly coarse segmentation (e.g., multiple materials merging into a single region or a single region being split up into multiple segmentation classes) can degrade the final reconstruction. This performance decrease is however not severe and thus for all conducted experiments a total of 100 novel render viewpoints are leveraged to create a robust segmentation.

\section{Additional Results}
\label{sec:add_res}

\begin{figure*}
  \centering
  \includegraphics[width=1\linewidth]{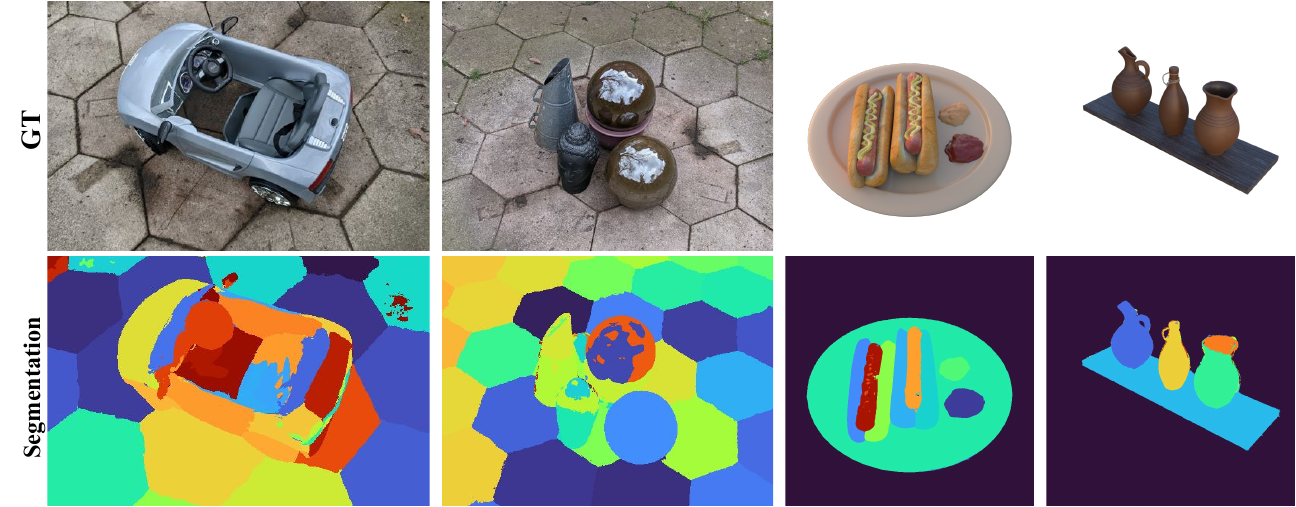}
  \caption{\textbf{Reconstructed segmentation maps generated by our iterative lifting procedure}. The first row shows the reference ground truth images, while the second row displays our rendered segmentation maps. For each scene, we render 100 novel viewpoints in an orbital trajectory around the object to perform iterative segmentation lifting and Gaussian-object merging. Despite the challenging setting of using only 8 input training views, our method produces coherent and largely accurate segmentations, with only minor missegmentation in a few regions.}
  \label{fig:supp_segmentation}
  \vspace{-0.5em}
\end{figure*}

We present additional quantitative results on the TensorIR~\cite{Jin2023TensoIR} dataset in \autoref{tab:tensorIR_supp} to further validate the effectiveness of our method. Even on diffuse objects, our method surpasses existing Gaussian-based inverse rendering methods and demonstrates its effectiveness especially in relighting. \\
Furthermore, \autoref{fig:supp_segmentation} shows reconstructed segmentation maps for several representative scenes, highlighting the structural coherence of our lifted segmentation across viewpoints.

\begin{table*}[h!]
\centering
\caption{\textbf{Quantitative evaluation on the synthetic TensorIR~\cite{Jin2023TensoIR} dataset.} GAINS estimates better NVS and relighting then baselines leading to better relighting performance, while competing with GI-GS on albedo estimations.
(\colorbox{1}{\raisebox{0pt}[1ex][0ex]{Red}} = best, 
    \colorbox{2}{\raisebox{0pt}[1ex][0ex]{Orange}} = 2nd best, 
    and \colorbox{3}{\raisebox{0pt}[1ex][0ex]{Yellow}} = 3rd best)
}
\label{tab:tensorIR_supp}
\begin{adjustbox}{max width=\textwidth,center}
\begin{tabular}{l | c c c | c c c | c c c |}
\multicolumn{10}{c}{\textbf{TensorIR~\cite{Jin2023TensoIR} dataset (8 views)}} \\
\hline
\multirow{2}{*}{\textbf{Methods}} &
\multicolumn{3}{c|}{\textbf{NVS}} &
\multicolumn{3}{c|}{\textbf{Albedo}} &
\multicolumn{3}{c|}{\textbf{Relight}} \\
\cline{2-10}
 &
\textbf{PSNR} $\uparrow$ & \textbf{SSIM} $\uparrow$ & \textbf{LPIPS} $\downarrow$ &
\textbf{PSNR} $\uparrow$ & \textbf{SSIM} $\uparrow$ & \textbf{LPIPS} $\downarrow$ &
\textbf{PSNR} $\uparrow$ & \textbf{SSIM} $\uparrow$ & \textbf{LPIPS} $\downarrow$ \\
\hline
Ref-Gaussian \cite{yao2025refGS} & \cellcolor{3}24.86 & \cellcolor{3}0.83 & \cellcolor{2}0.11 & \cellcolor{3}19.86 & \cellcolor{3}0.70 & \cellcolor{3}0.21 & \cellcolor{3}25.20 & \cellcolor{3}0.71 & \cellcolor{2}0.12 \\
GI-GS \cite{chen2025gigs} & \cellcolor{2}28.01 & \cellcolor{2}0.89 & \cellcolor{2}0.11 & \cellcolor{1}28.96 & \cellcolor{2}0.90 & \cellcolor{2}0.13 & \cellcolor{2}25.40 & \cellcolor{2}0.85 & \cellcolor{3}0.13 \\
Ours & \cellcolor{1}28.34 & \cellcolor{1}0.91 & \cellcolor{1}0.10 & \cellcolor{2}27.10 & \cellcolor{1}0.91 & \cellcolor{1}0.12 & \cellcolor{1}27.70 & \cellcolor{1}0.90 & \cellcolor{1}0.11 \\
\hline
\end{tabular}
\end{adjustbox}
\end{table*}

\end{document}